\documentclass{bmvc2k}
\usepackage{booktabs}
\usepackage{caption}
\usepackage{subcaption}
\usepackage{soul}
\usepackage{graphicx}
\usepackage{amsmath}
\usepackage{amssymb}
\usepackage{booktabs}
\usepackage{multirow}
\usepackage{bm}
\usepackage{comment}
\usepackage{xr-hyper}
\usepackage[capitalize]{cleveref}
\crefname{section}{Sec.}{Secs.}
\Crefname{section}{Section}{Sections}
\Crefname{table}{Table}{Tables}
\crefname{table}{Tab.}{Tabs.}

\usepackage{xcolor}

\definecolor{ai-teacher}{RGB}{0, 168, 238}
\definecolor{ai-student}{RGB}{0, 168, 70}
\definecolor{CAM}{RGB}{217, 130, 43}
\definecolor{RISE}{RGB}{190, 96, 235}

\begin{document}

\title{Teaching AI to Teach: \\Leveraging Limited Human Salience Data Into Unlimited Saliency-Based Training}

\addauthor{Colton R. Crum}{ccrum@nd.edu}{1}
\addauthor{Aidan Boyd}{aboyd3@nd.edu}{1}
\addauthor{Kevin W. Bowyer}{kwb@nd.edu}{1}
\addauthor{Adam Czajka}{aczajka@nd.edu}{1}

\addinstitution{
 University of Notre Dame\\
 Notre Dame, IN 46556, USA
}

\runninghead{Crum, Boyd, Bowyer, Czajka}{Teaching AI to Teach}

\def\eg{\emph{e.g}\bmvaOneDot}
\def\ie{\emph{i.e}\bmvaOneDot}
\def\Eg{\emph{E.g}\bmvaOneDot}
\def\etal{\emph{et al}\bmvaOneDot}

\maketitle

Machine learning models have shown increased accuracy in classification tasks when the training process incorporates human perceptual information. 
However, a challenge in training human-guided models is the cost associated with collecting image annotations for human salience.
Collecting annotation data for all images in a large training set can be prohibitively expensive.
In this work, we utilize ``teacher'' models (trained on a small amount of human-annotated data) to annotate additional data by means of teacher models' saliency maps.
Then, ``student'' models are trained using the larger amount of annotated training data.
This approach makes it possible to supplement a limited number of \emph{human-supplied} annotations with an arbitrarily large number of \emph{model-generated} image annotations. 
We compare the accuracy achieved by our teacher-student training paradigm with (1) training using all available human salience annotations, and (2) using all available training data without human salience annotations.
We use synthetic face detection and fake iris detection as example challenging problems, and report results across four model architectures (DenseNet, ResNet, Xception, and Inception), and two saliency estimation methods (CAM and RISE).
Results show that our teacher-student training paradigm results in models that significantly exceed the performance of both baselines, demonstrating that our approach can usefully leverage a small amount of human  annotations to generate salience maps for an arbitrary amount of additional training data.

\section{Introduction}
\label{sec:intro}

Computer vision architectures often take inspiration from brain physiology, mental models, and attention mechanisms, which can be incorporated into the training of models in many different ways. A common way to train human-guided models is through saliency-based training, which has shown to (a) generalize better to new data, which is vital in open-set recognition where not all classes are known, (b) speed up training time by using less samples that contain more meaningful information (data + associated saliency), (c) increase the model’s focus on class features, limiting sensitivity to features accidentally correlated with class labels, and (d) produce more human explainable outputs.
One limitation of human-guided models can be the high cost to acquire human perception-related information, such as image annotations.
One potential solution to address this limitation is to build models that are capable of generating human-like saliency maps to annotate new data used to train subsequent models.

\begin{figure}[!ht]
\includegraphics[width=\linewidth]{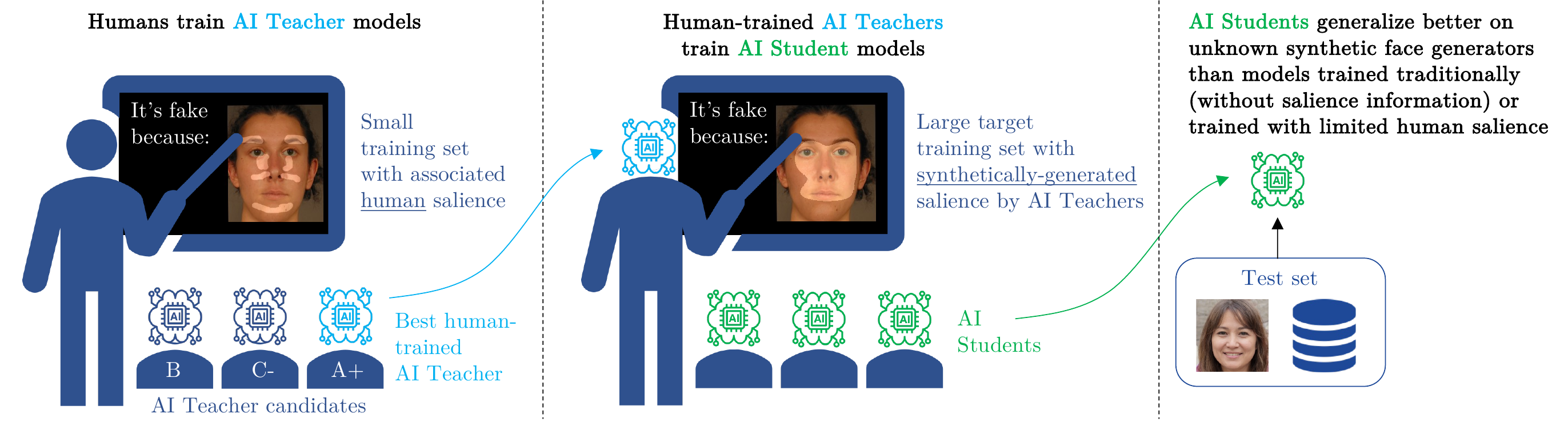}
\caption{This work explores training {\it AI Teachers} -- models first trained in a human-guided way -- to train future {\it AI Student} models. AI Teachers are selected from models that have the highest Area Under the ROC Curve (AUC) on the validation set and provide saliency for larger, unannotated training data, eventually used to train AI Students. With this approach, always-limited human annotations are efficiently leveraged to provide salience data to an unlimited number of training samples.}
\label{fig:teaser}
\end{figure}

We explore a training framework which first uses the available human-salience data to train an {\it AI Teacher} model, which is then used to generate saliency maps, similar to human saliency, for large amounts of additional training data (see Fig. \ref{fig:teaser}).
The {\it AI Student} model is then trained using the training data annotated by the AI Teacher. The AI Teacher's saliency maps can be generated using white-box approaches (e.g., CAM \cite{cam}), or black-box approaches that rely on perturbing the input image and observing the effect on the output (\eg, RISE \cite{petsiuk2018rise}).

We experiment with our teacher-student framework for synthetic face detection using the CYBORG human-guided training paradigm \cite{boyd2021cyborg}. 
However, our approach is applicable to any task for which humans can provide salience, and we present its viability also for fake iris detection. 

We show that the performance of AI Student models, trained by the human-taught AI Teacher, surpasses the performance of both (a) models trained with limited human salience (Baseline 1), and (b) models trained without human salience but on large training data (Baseline 2). Thus, the proposed approach provides a means to efficiently convert a small amount of human-provided salience data into a large amount of effective human-like saliency. Our framework allows for increased data diversity and new information for each training sample, which exceeds performance rather than simply adding more data. Results in this paper are organized around the following {\bf research questions}:
\begin{itemize}
\addtolength\itemsep{-2mm}

    \item \textbf{RQ1:} Which type of training produces better AI Teacher models: human-guided or purely data-driven? We consider four CNN architectures with the same architecture for teacher and student models. (\emph{Sec. \ref{sec:RQ1} and Tab. \ref{tab:face-results})}

    \item \textbf{RQ2:} Can the top-performing AI Teacher model improve the performance of AI Student models across different CNN architectures? (\emph{Sec. \ref{sec:RQ2} and Fig.  \ref{fig:best-worst-experiment}})

    \item \textbf{RQ3:} What are the potential performance benefits of the teacher-student training paradigm over the baselines? (\emph{Sec. \ref{sec:RQ3} and Tab. \ref{tab:face-optimal}})
    
    \item \textbf{RQ4:} Can this training approach be applied to domains beyond synthetic face detection? (\emph{Sec. \ref{sec:RQ4}, Tab. \ref{tab:iris-optimal}}.)
    
\end{itemize}

\section{Related Work}

\paragraph{Estimating Model Salience} \label{sec:salience-methods} Access to models' internal data (feature maps, gradients, weights) simplifies building saliency estimation methods.
Class Activation Mapping (CAM) is the most popular approach to estimate salience of white-box models \cite{cam}. CAM works by making a forward pass through the model to get the activations of the last convolutional layer, which are weighted into a heat map.
A potential downside of CAM is low resolution of the resulting visualization; \eg, $7\times7$ for DenseNet.
Recent advances
such as Grad-CAM
\cite{gradcam}, Grad-CAM++ \cite{gradcam++}, HiResCAM \cite{hirescam}, Score-CAM \cite{scorecam}, Ablation-CAM \cite{ablationcam}, or Eigen-CAM \cite{eigencam} aim to provide more detailed saliency estimations, but require more computational resources.

In case of black-box models, dominant methods rely on a simple idea of randomly perturbing input regions and observing the impact on the output. Random Input Sampling Explanation (RISE) \cite{petsiuk2018rise} is one such method, in which a weighted average (where the weights serves as the ``confidence'' scores) is used to generate a full-sized salience map. Black-box approaches, such as RISE, have two main benefits; (1) they require no information from inside the model, and (2) the resolution of generated salience may be as high as that of the input image, whereas CAMs are limited to the spatial dimensions of the last convolutional layer. 
Recent work on black-box explainers include methods of evaluating their usefulness for humans \cite{Carmichael_ArXiv_2023} and increasing their robustness against adversarial attacks \cite{Carmichael_AAAI_2023}.

The above mentioned techniques are part of the broader and  dynamic ``eXplainable AI'' (XAI) area \cite{Saleem_Neurocomputing_2022}. In this work, we use CAM and RISE to compare their usefulness in generating salience of teacher models, since saliency methods can be architecture-specific and may impact the performance of the Teacher-Student training paradigm.

\paragraph{Human Salience-Guided Model Training} Incorporating human perceptual capabilities into the model training is non-trivial, and may involve human-sourced information in various forms: image/video annotations \cite{boyd2021cyborg}, eye-tracking \cite{boyd2023human,Czajka_WACV_2019}, reaction times \cite{huang2022measuring}, or even games \cite{linsley2018learning}. Successful ways of incorporating human-collected information into training include adding specialized components to the loss functions \cite{boyd2021cyborg, huang2022measuring}, augmenting training data \cite{boyd2022human}, and introduction of human perception-based regularization \cite{huang2022measuring, dulay2022using}. Specifically, ConveYing Brain Oversight to Raise Generalization (CYBORG) training strategy \cite{boyd2021cyborg} combines both human and model's salience into the loss function by penalizing the divergence of the model's CAM from the human salience provided as image annotations. Application of the CYBORG loss function increased the performance in synthetic face detection across four, out-of-the-box CNN architectures using only 1,821 training samples with associated human annotations, compared to models trained with cross-entropy loss.

Although human salience-based training has been successfully implemented (\eg CYBORG), to our knowledge the approaches that would enable more effective use of human annotations, and thus scale the human-guided training, have not yet been explored. We find this an important and essential research direction in the future of training human-guided models for a number of applications. This paper shows how to create ``proxy'' models, which by producing human-like salience for any input, allow human-inspired annotations to train models on without additional cost.

\section{Methodology}

In this section, we first describe the two baselines that we compare our training paradigm against. Secondly, we describe our dataset splits to train AI Teachers (TAIT), train AI Students (TAIS) and evaluate AI Students (EAIS). We then describe our method for creating representative human-guided AI Teachers, including experimental design parameters and teacher model selection. Finally, we describe the performance metrics used to evaluate research questions RQ1-RQ4.

\subsection{Baseline Models}
\label{sec:baselines}
We benchmark our Teacher-Student training paradigm against two baselines.
Baseline 1 is to simply train human-guided models using the available human annotations, which was previously proposed in \cite{boyd2021cyborg}. Within the naming conventions of this paper, Baseline 1 can be thought of as simply using the teacher models on the test set, without training and using any student models. Baseline 2 is to train traditional  (no saliency) models on all available training data. These two baselines represent contrasting viewpoints in achieving optimal model performance: (a) giving the model human-guided information on ``where to look'' in order to solve the task (Baseline 1), or (b) giving the model a large amount of data
to train on (Baseline 2). 
For some tasks and domains, Baseline 1 or 2 may achieve the desired performance. However, for Baseline 1, the vast majority of training data remains un-annotated and is not used.
And for Baseline 2, the hope is that ``enough'' training data has been used to train an optimal model. 
Our approach is a strategic blend of using human-generated salience for whatever fraction of training data it is available, and using model-generated salience for all remaining training data.

\subsection{Datasets}

\begin{figure*}[!h]
    \centering
    \includegraphics[width=\linewidth]{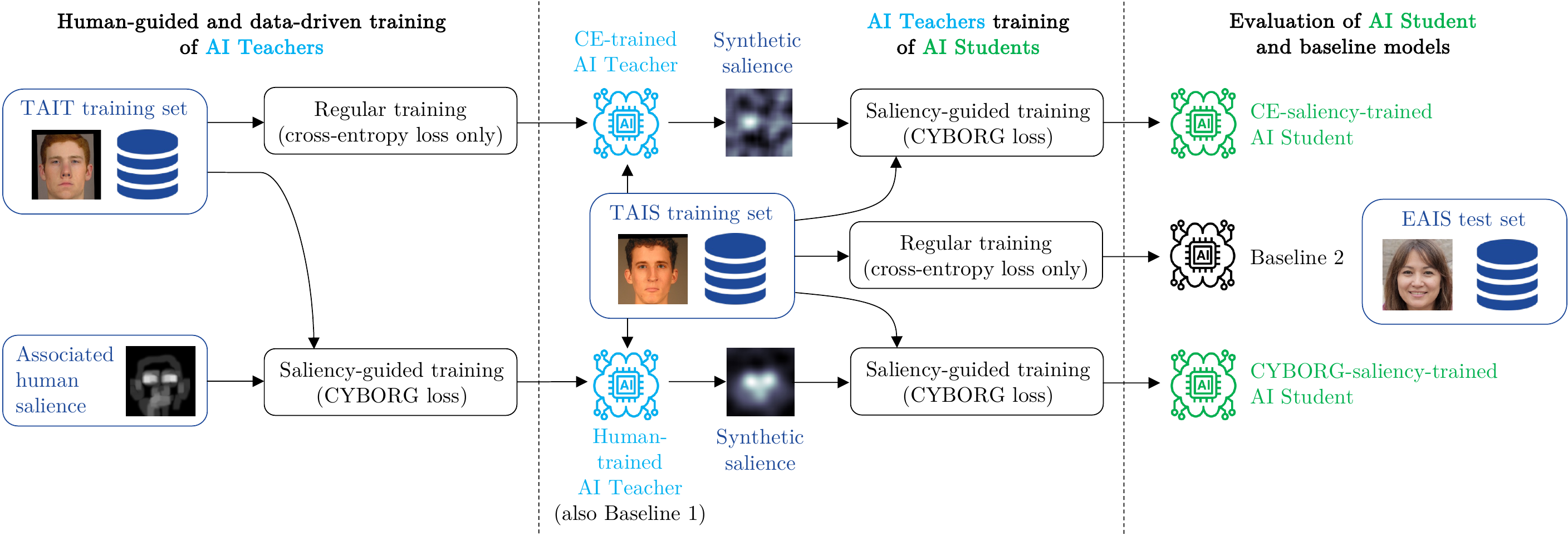}
      \vskip2mm
    \caption{Detailed diagram of data and saliency usage.}
    \label{fig:data_usage}
\end{figure*}
This section presents three dataset splits that play essential roles in the proposed pipeline (cf. Fig. \ref{fig:data_usage}). First, the \textcolor{ai-teacher}{AI Teachers} are trained on a (potentially small) training dataset with saliency maps sourced from human annotations. Next, the best \textcolor{ai-teacher}{AI Teacher} is used to generate synthetic (yet human-like) saliency maps for samples in target (potentially large) training dataset. This creates \textcolor{ai-student}{AI Students}, which generalize better to unknown samples in the test set, compared to (a) student models trained without any saliency, and (b) student models trained with synthetic saliency, but generated by teacher models trained without human perceptual inputs. 

To evaluate our training framework, we use the task of synthetic face detection. Selected results are repeated also for iris presentation attack detection (PAD) to address the generalization capability of the proposed approach across domains.

\paragraph{Dataset for Training AI Teachers (TAIT)} 
These are the subsets of the overall training sets that have human-provided annotations of salience, and are used to {\bf T}rain the {\bf AI} {\bf T}eachers.
 For the human-guided training of AI Teachers in the {\bf synthetic face detection} task, we use the same dataset and human annotations as introduced in \cite{boyd2021cyborg}\footnote{The authors of this paper would like to thank the authors of \cite{boyd2021cyborg} for sharing their data with us.}. The training set consists of 919 authentic and 902 synthetic face images with annotations of regions selected by 363 humans (recruited via Amazon Mechanical Turk) as important to them judging the authenticity of a given face image. Only annotations for correctly classified image pairs are used for the training set. The validation set is composed of 10,000 authentic faces (sampled from FRGC-Subset~\cite{Phillips_IVC_2017}) and 10,000 synthetic faces (sampled from SREFI \cite{Banerjee_IJCB_2017} and from StyleGAN2-generated images at \emph{thispersondoesnotexist.com}). All images were pre-processed using img2pose \cite{albiero2021img2pose}, resized to $224\times224$, and cropped to ensure the face is in full view. Subjects were presented with a pair of face images, one authentic and one synthetic, and asked an alternating prompt of which face is real (fake). After answering the question, subjects used their cursor to highlight regions of the selected face that support their decision. The human salience maps were cropped and resized to $224\times224$ in the same way as the image data to match the corresponding input images. 
 
 For training {\bf iris PAD} AI Teachers, we used 765 samples annotated by humans, offered with \cite{boyd2022human}, including bona fide irises, and seven spoof attack types (artificial, textured contact lens, post mortem, paper print outs, synthetic, diseased, textured contact lens \& printed). Only correctly classified samples were used in model training.

 The TAIT image sets, without the human-salience heatmaps, are also used to train models with cross-entropy loss in order to answer Research Question 1.
 
\paragraph{Datasets for Training AI Students (TAIS)}
Larger training sets, for which no human annotations of salience are available, were used to {\bf T}rain the {\bf AI} {\bf S}tudents. For \textbf{synthetic face detection} task, the dataset was collected from the same sources as the TAIT (FRGC, SREFI and StyleGAN2; see example images added to the supplementary materials). 
This resulted in a TAIS dataset six times larger than, and image-disjoint from, the smaller TAIT dataset for synthetic face detection task. The TAIS dataset for \textbf{iris PAD} task was collected from the same sources as TAIT for iris PAD, and is certainly image-disjoint. However, due to a more challenging scenario of collecting physical iris spoofs, we kept the size of TAIS similar as the size of TAIT except so that classes (live / spoof) could be completely balanced (764 samples).

Both TAIS datasets have no associated human saliency for the images. 
Instead, salient regions for each image are given by a salience map (either either CAM- or RISE-based) generated from the AI Teacher model that was trained using TAIT.

\paragraph{Datasets for Evaluating AI Students (EAIS)} 
In conventional machine learning terms, these are the test sets. However, we are careful to make this distinction as the teacher models are completely withheld from the test set, and only the student models are evaluated on this set. Instead, teacher models are assessed by their performance on the validation set (Sec. \ref{sec:generation_of_ai_teachers} discusses the teacher model selection process in depth). For \textbf{synthetic face detection} task, the EAIS set contains (a) 600,000 synthetic face images, evenly sampled from six GAN architectures (ProGAN~\cite{karras2017progressive}, StarGANv2 \cite{stargan}, StyleGAN \cite{styleGAN}, StyleGAN2 \cite{styleGAN2}, StyleGAN2-ADA \cite{syleGAN2-ADA}, and StyleGAN3\cite{styleGAN3}; samples are presented in supplementary materials), and (b) 100,000 authentic face images: 70,000 from FFHQ and 30,000 from CelebA-HQ \cite{Celeb}. ProGAN and StarGANv2 were trained using CelebA-HQ, whereas the rest of the GAN generators (StyleGAN, StyleGAN2, StyleGAN2-ADA, and StyleGAN3) were trained using FFHQ. For the {\bf iris PAD} task, the test set is comprised of 12,432 samples across six categories (live, artificial, texted contact lenses, display, post mortem, and paper print outs), which is identical to the test set used in the LivDet-Iris-2020 competition benchmark \cite{das2020iris}.

\subsection{Performance Metrics}

In an effort to benchmark our results against the most recent human saliency-based training, we first conducted experiments using the exact same dataset sources, model backbones, and experimental environment as in \cite{boyd2021cyborg}. To assess the uncertainty related to random training seeds, we trained 10 models for each discussed dataset-model configuration. Area Under the ROC Curve (AUC) is used to compare the performance of the models.

\subsection{Generation of Human-Guided AI Teachers}
\label{sec:generation_of_ai_teachers}

\paragraph{Saliency-based Model Training} Our framework for teaching AI Teachers begins by first training 10 models on the TAIT dataset using human annotations and the CYBORG loss, which simultaneously maximizes the classification performance, and closeness of the model and human saliency maps \cite{boyd2021cyborg}:
\begin{equation}
\label{equation:cyborg}
\mathcal{L} = \frac{1}{K}\sum_{k=1}^K\sum_{c=1}^{C}\bm{1}_{y_k \in \mathcal{C}_c} \Bigg[\underbrace{(1-\alpha)\|\textbf{s}_k^{\text{(teacher)}}  - \textbf{s}_k^{\text{(model)}}\|^2}_{\text{teacher saliency loss component}} -\underbrace{\alpha\log p_{\text{model}}\big(y_k \in \mathcal{C}_c\big)}_{\text{classification loss component}}\Bigg]
\end{equation}
\noindent
where $\|\cdot\|$ is the $\ell_2$ norm, $y_k$ is a class label for the $k$-th sample, $\bm{1}$ equals to $1$ when $y_k \in \mathcal{C}_c$ (and equals to 0 otherwise), $C$ is the number of classes, $K$ is a batch size, $\alpha=0.5$ is a trade-off parameter weighting teacher- and model-based saliency maps. The $\textbf{s}_k^{\text{(teacher)}}$ is the salience generated by the teacher (\ie by a human in case of teaching the AI Teacher models, or by the AI Teacher in case of teaching the AI Students) for the $k$-th sample. The $\textbf{s}_k^{(\text{model})}$ is the model saliency estimated by weighting all features maps in the last convolutional layer using weights in the last classification layer belonging to the predicted class $\mathcal{C}_c$. We follow \cite{boyd2021cyborg} and normalize both  $\textbf{s}_k^{(\text{model})}$ and $\textbf{s}_k^{(\text{teacher})}$ to the range of $\langle 0,1\rangle$.

Next, the model with the highest AUC on the validation part of the TAIT dataset is selected as the AI Teacher. The selected teacher model then generates saliency (using either RISE or CAM approach) on the larger unannotated TAIS training set. Finally, 10 subsequent AI Students are trained on the TAIS dataset with the associated AI Teacher-generated salience maps using CYBORG loss. 

\paragraph{Model Architectures} We used four out-of-the-box architectures across all experiments: DenseNet121 \cite{huang2017densely}, ResNet50 \cite{resnet}, Xception \cite{chollet2017xception}, and Inception v3 \cite{szegedy2016rethinking}. All model weights were instantiated from the pre-trained ImageNet weights. All models were trained using Stochastic Gradient Descent (SGD) for maximum 50 epochs, with learning rate of 0.005, modified by a factor of 0.1 every 12 epochs. The initial teacher saliency and model saliency components in the human-guided (CYBORG) loss were given equal weighting, \ie $\alpha=0.5$ in Eq. \eqref{equation:cyborg}, as in \cite{boyd2021cyborg} and \cite{boyd2022human}. Optimal student model configurations were achieved by lowering $\alpha=0.01$ (the exploration of the weighting parameter $\alpha$ is detailed in Section \ref{sec:RQ1}).

\section{Results}
\paragraph{Answering RQ1: Which type of training makes better AI Teacher models: human guided or purely data driven?}
\label{sec:RQ1}

\begin{table}[!tb]
\centering
\caption{Mean Area Under the Curve (AUC) for synthetic face detection task solved by all variants of AI Students. Ten models were trained for each variant and standard deviations are given. Only one architecture (Inception) did not benefit from the salience generated by AI Teachers taught initially by humans. Best results for each model architecture are {\bf bolded}, and better type of saliency is color coded: \textcolor{RISE}{RISE}, and \textcolor{CAM}{CAM}.}
\vskip5mm
\label{tab:face-results}
\begingroup
\setlength{\tabcolsep}{2pt}
\renewcommand{\arraystretch}{1}
\small
\begin{tabular}{ccccc}
\toprule
{\bf How the AI Teachers}  & \multicolumn{4}{c}{\bf Mean AUC on the EAIS data} \\\cline{2-5}
{\bf were trained on TAIT data}  & {\bf DenseNet} & {\bf ResNet} & {\bf Xception} & {\bf Inception}\\

\midrule

 Without human salience  & \textcolor{RISE}{0.591} \textcolor{RISE}{$\pm$}\textcolor{RISE}{0.036} & \textcolor{RISE}{0.601}\textcolor{RISE}{$\pm$}\textcolor{RISE}{0.019} & \textcolor{CAM}{0.694}\textcolor{CAM}{$\pm$}\textcolor{CAM}{0.011} & \textcolor{RISE}{\textbf{0.645}}\textcolor{RISE}{$\pm$}\textcolor{RISE}{\textbf{0.020}} \\
  \cline{1-5}
 With human salience  & \textcolor{RISE}{\textbf{0.696}}\textcolor{RISE}{$\pm$}\textcolor{RISE}{\textbf{0.016}} & \textcolor{RISE}{\textbf{0.634}}\textcolor{RISE}{$\pm$}\textcolor{RISE}{\textbf{0.021}} & \textcolor{CAM}{\textbf{0.722}}\textcolor{CAM}{$\pm$}\textcolor{CAM}{\textbf{0.011}} & \textcolor{CAM}{0.617}\textcolor{CAM}{$\pm$}\textcolor{CAM}{0.040} \\
   \bottomrule
 \end{tabular}
 \endgroup
 \end{table}
To fairly assess the value of using human salience in training AI Teachers, we first taught 10 of such models using TAIT data with human saliency. In order to answer research question RQ1, we additionally trained another 10 AI Teachers with only cross-entropy loss (``CE-trained AI Teacher'' in Fig. \ref{fig:data_usage}). This is to investigate if human annotations are at all needed at any step of the entire framework. For both tasks, three out of four AI Student model architectures benefited from AI Teachers being trained with human annotations as opposed to being trained without human salience, as seen in Tab. \ref{tab:face-results} for synthetic face detection task. For the one AI Student model architecture that did not benefit from AI Teachers being trained with human salience (Inception), we believe the standard deviations indicate these differences weren't statistically significant. More specifically, we believe this result is due to the selection of a poor Teacher model. The Teacher-Student training paradigm selects the highest-performing model on the validation set as the Teacher. However, this may not always generate the best salience due to overfitting, or latching onto spurious features despite presence of human salience. This is illustrated in Fig. S4 in the supplementary materials, which shows that the selected Inception-based Teacher model failed to focus on important regions of the input image (see specifically col (e) row (ii) in that figure). With Inception Teacher's saliency maps unfocused on the wrong features, the AI Student's performance will inevitably suffer.

Thus, {\bf the answer to RQ1 is affirmative: effective student models benefit from being trained with AI Teachers trained with human-salience, compared to AI Students taught by teachers not exposed to human salience}.
\paragraph{Answering RQ2: Can the top performing AI Teacher improve the performance of AI Students across different CNN architectures?}
\label{sec:RQ2}
\begin{figure}[!htb]
    \centering
      \begin{subfigure}[b]{0.263\linewidth}
          \centering
          \includegraphics[width=\linewidth]{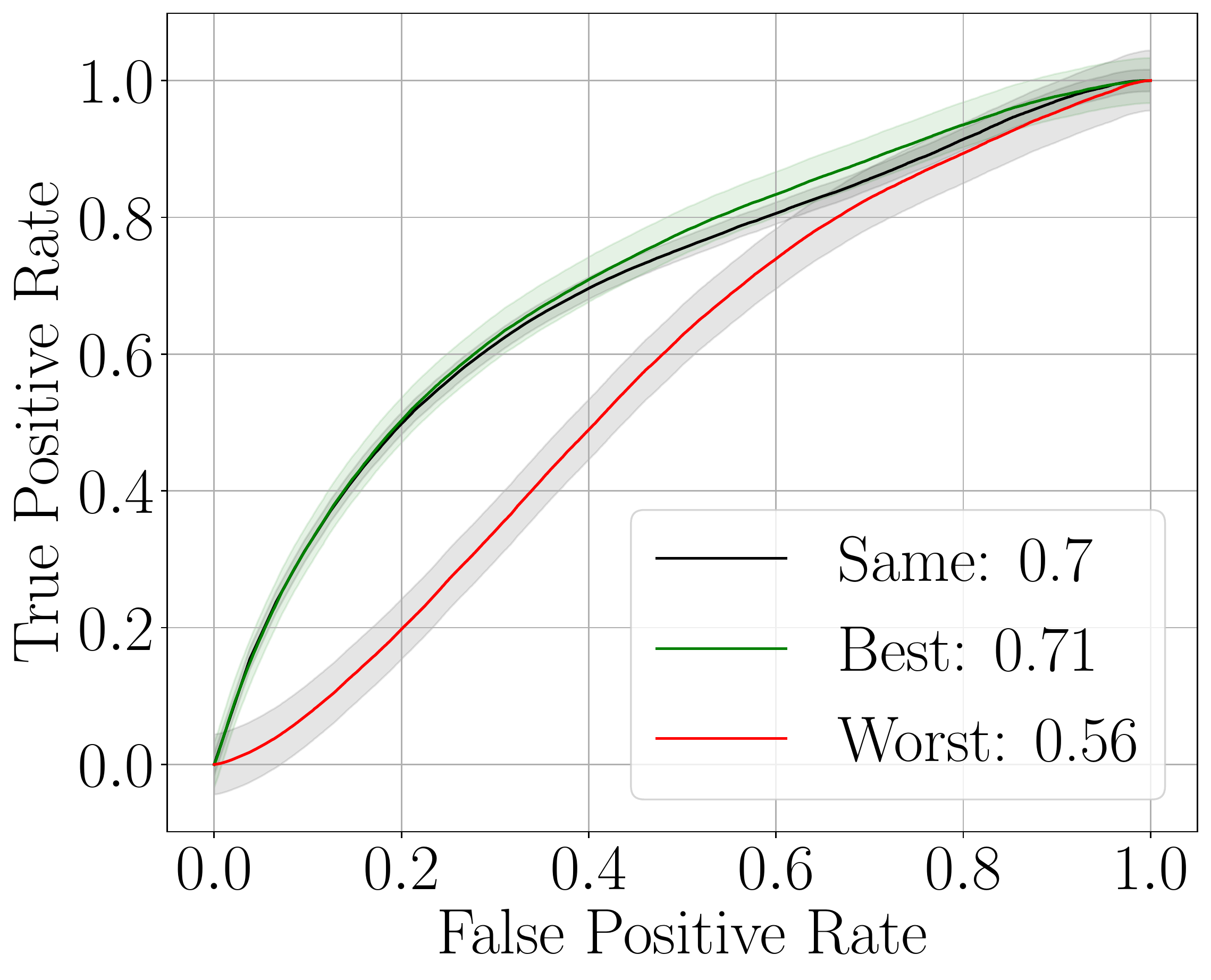}
          \caption{DenseNet}
      \end{subfigure}\hfill
      \begin{subfigure}[b]{0.23\linewidth}
          \centering
          \includegraphics[width=\linewidth]{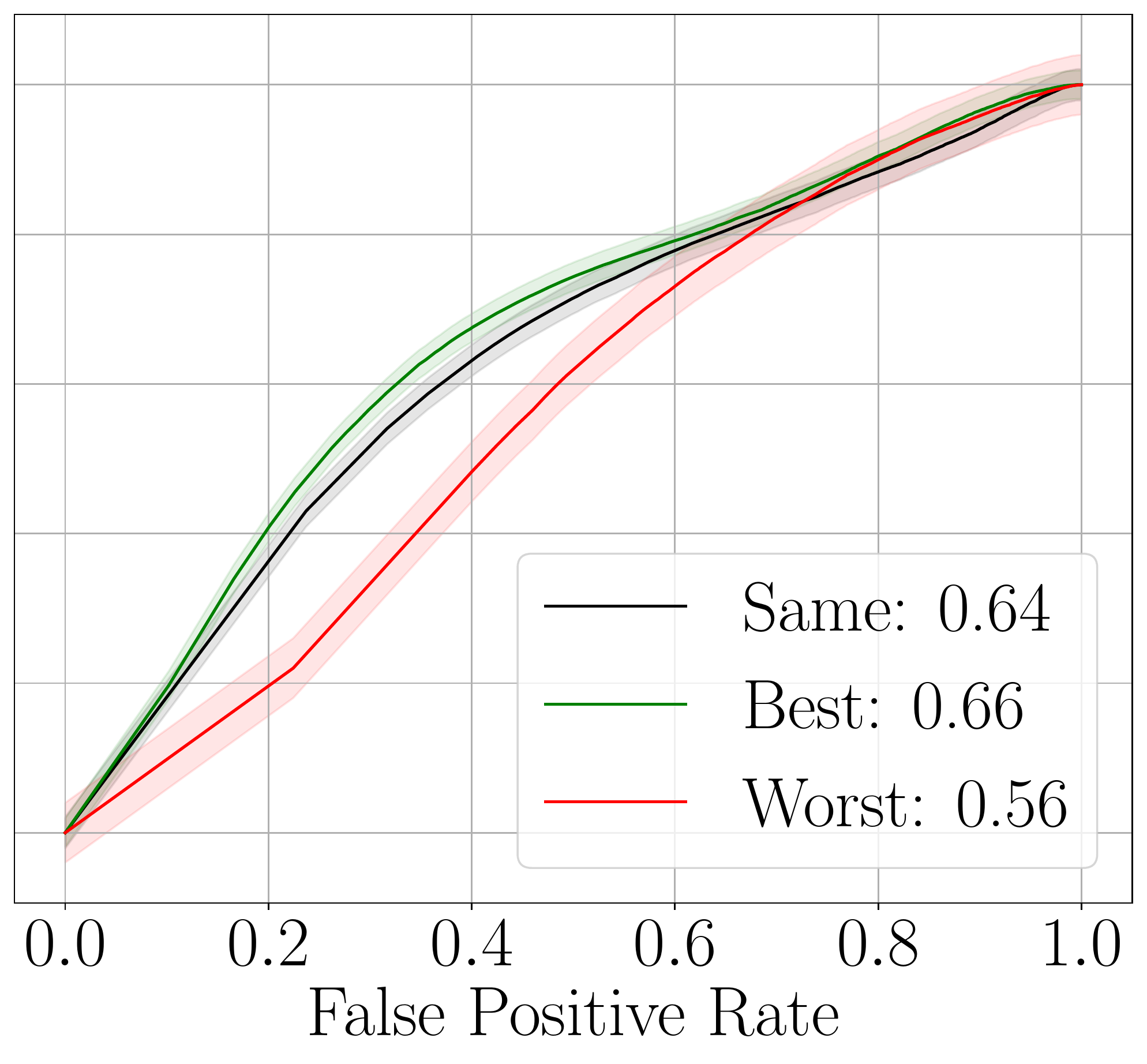}
          \caption{ResNet}
      \end{subfigure}\hfill
      \begin{subfigure}[b]{0.23\linewidth}
          \centering
          \includegraphics[width=\linewidth]{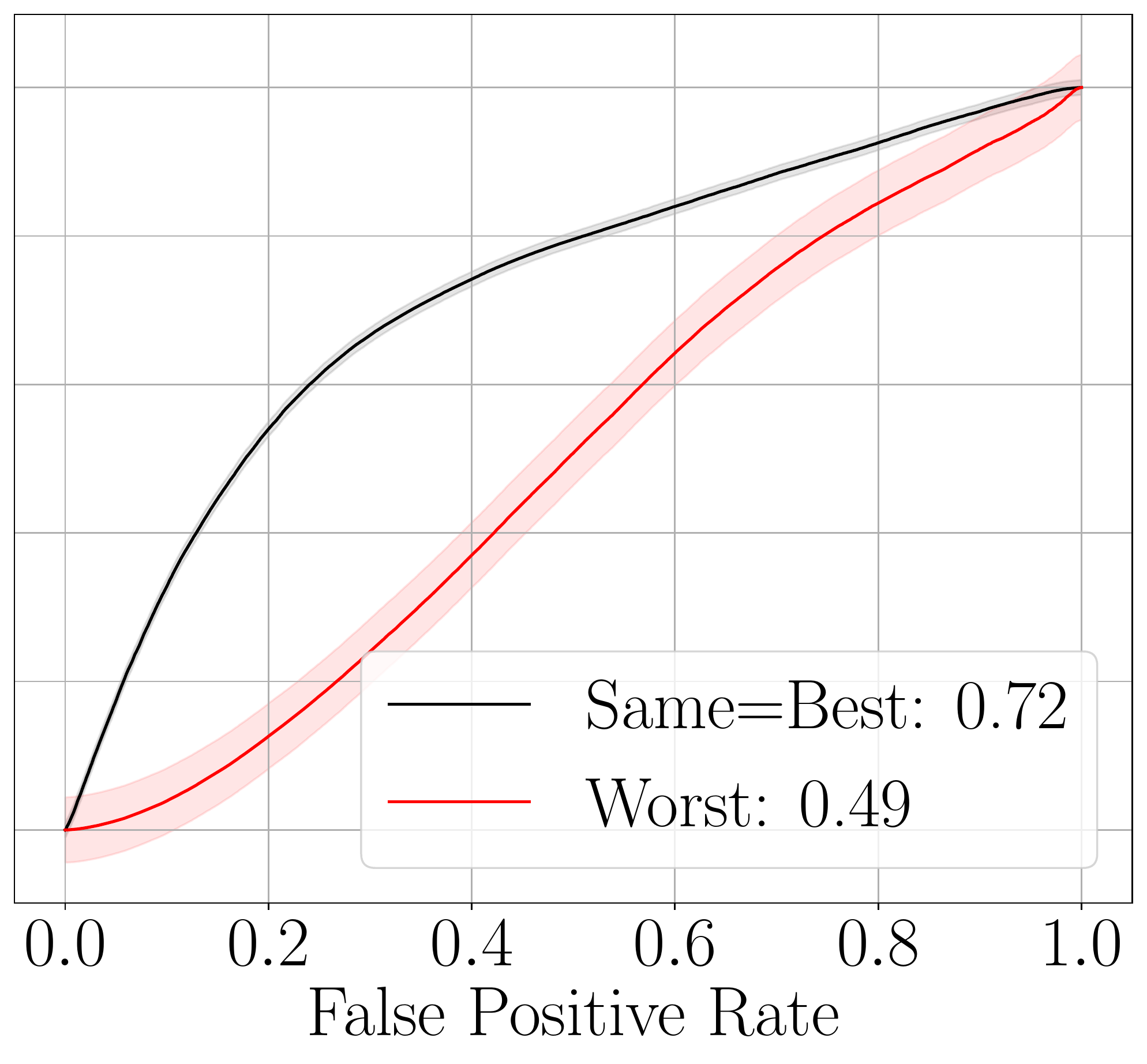}
          \caption{Xception}
      \end{subfigure}\hfill
      \begin{subfigure}[b]{0.23\linewidth}
          \centering
          \includegraphics[width=\linewidth]{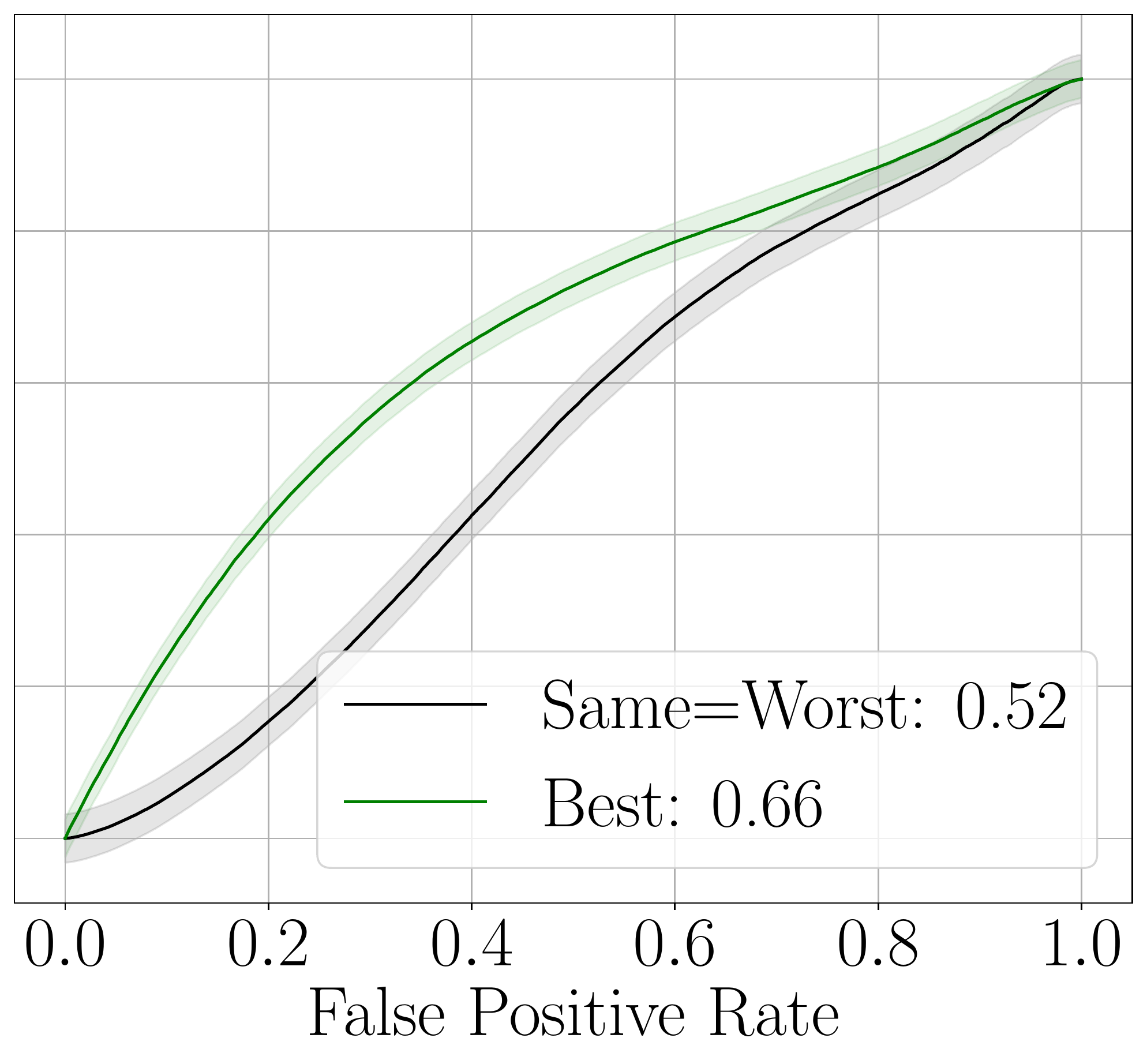}
          \caption{Inception}
      \end{subfigure}
    \caption{Mean ROC curves, along with bands representing standard deviations across 10 independent train-test runs, illustrating how the performance of the AI Teacher (and resulting salience generated by that teacher model) impacts the AI Student's performance. ``Same'' indicates that both AI Teacher and AI Student shared the same architecture. For example, ``Same'' for DenseNet indicates that the student model was trained using saliency generated by the DenseNet-based teacher model. ``Best'' means that the student model was trained with saliency generated by the best AI Teacher, possibly with a different architecture. For comparison, the ``Worst'' means that the student model was trained with saliency generated by the worst AI Teacher. For ROCs denoted as ``Same = Best'' or ``Same = Worst'', there was no difference in performance for these options, so we keep one ROC curve.}
    \label{fig:best-worst-experiment}
\end{figure}

One of the most profound insights is that AI Teacher's saliency is transferable across different CNN architectures. To explore this, we applied the AI Teacher which taught the best-performing student model (Xception-based teacher with CAM-based saliency, AUC=0.722 in Tab. \ref{tab:face-results}), to train another student models, but based on different architectures (DenseNet, ResNet, and Inception) than the AI teacher. Every model achieved better performance using Xceptions's CAM-based saliency instead of the saliency generated by a teacher sharing the same architecture as a student. As a sanity check, we then performed the opposite experiment: applying the AI Teacher that resulted in the worst-performing student models (in this case Inception-based teacher model with RISE-based saliency, AUC=0.508) to train the student model (again, based on a different architecture than the AI Teacher: DenseNet, ResNet and Xception). Every model that used Inception's saliency decreased its performance significantly. Fig. \ref{fig:best-worst-experiment} illustrates the results via ROC curves. From these experiments, we can conclude that the best teacher models are learning more salient features from the data, and passing that information along effectively to student models. Thus, the {\bf answer to RQ2 is affirmative: the top performing AI Teacher can improve the performance of AI Students and does not need to have the same experimental parameters or even architecture to convey saliency-related information efficiently to future student models}.

\paragraph{Answering RQ3: What are the potential performance benefits of the teacher-student training paradigm over the baselines?}
\label{sec:RQ3}
\begin{table}[!htb]
\centering
\caption{Area Under the Curve (AUC, mean $\pm$ std) achieved by the baselines and the optimal student model in synthetic face detection task. Optimal AI Student configurations were achieved by training the model using the optimal AI Teacher's configuration (Xception + CAM) with $\alpha=0.01$ (\ie encouraging the model to focus on salience instead of class label).}
\vskip5mm
\label{tab:face-optimal}
\small
\begin{tabular}{lccc}
\toprule
\textbf{Model} & \textbf{Baseline 1} & \textbf{Baseline 2} & \textbf{Optimal AI Student}\\
& (small set, entire & (larger set, no & (this paper: large set, optimal\\
& human salience available) &human salience) &  use of human salience) \\
\midrule
\textbf{DenseNet}  & 0.633 $\pm$ 0.04 & 0.629 $\pm$ 0.039 &\textbf{0.767 $\pm$ 0.020} \\
\textbf{ResNet} & 0.612 $\pm$ 0.05 & 0.555 $\pm$ 0.061& \textbf{0.718 $\pm$ 0.012} \\ 
\textbf{Xception} & 0.730 $\pm$ 0.02 & 0.586 $\pm$ 0.074 & \textbf{0.743 $\pm$ 0.005} \\ 
\textbf{Inception} & 0.679 $\pm$ 0.03 & 0.610 $\pm$ 0.035 & \textbf{0.746 $\pm$ 0.019} \\
\bottomrule
\end{tabular}
\end{table}

In answering this research question, we build upon the insights found from the previous two research questions in order to maximize the full capabilities of the proposed training paradigm. We explore increasing the performance of teacher-student training by: (1) using the ``best'' teacher model's saliency (conclusion from answering RQ1: teachers trained using human-salience teach better student models \& conclusion from answering RQ2: Xception's teacher using CAM saliency is best to teach students detecting synthetic faces), and (2) training the student models to ``look'' more aggressively at the teacher's saliency maps by lowering $\alpha$ in Eq. \eqref{equation:cyborg} during training to near zero ($\alpha=0.01$).
In addition to using the optimal teacher's saliency maps, once the AI Students have a more accurate map of ``where to look,'' the classification (cross entropy-based) component of the loss becomes less important to the student models.
The results from these experiments are reflected in Tab. \ref{tab:face-optimal}. As illustrated, \textbf{this approach significantly boosted the performance across all CNN architectures, and the accuracy of optimal student models trained with the proposed approach surpassed the accuracy of the baseline models.}.

\paragraph{Answering RQ4: Can this training approach be applied to domains beyond synthetic face detection?}
\label{sec:RQ4}

\begin{table}[!tb]
\centering
\caption{Same as in Table \ref{tab:face-optimal}, except that results for iris PAD are shown. Optimal AI Students may have different CNN architectures than their teacher models, and may have ``aggressive'' ($\alpha=0.01$) or modest ($\alpha=0.50$) weighting towards using human saliency.}
\vskip3mm
\label{tab:iris-optimal}
\small
\begin{tabular}{lccc}
\toprule
\textbf{Model} & \textbf{Baseline 1} & \textbf{Baseline 2} & \textbf{Optimal AI Student}\\
& (small set with the entire & (large set, & (this paper: large set, optimal\\
& human salience available) & no human salience) &  use of human salience) \\
\midrule
\textbf{DenseNet}  & 0.920 $\pm$ 0.017 & 0.917 $\pm$ 0.017 & \textbf{0.950 $\pm$ 0.013} \\ 
\textbf{ResNet} & 0.854 $\pm$ 0.031 & 0.905 $\pm$ 0.013 & \textbf{0.920 $\pm$ 0.022} \\ 
\textbf{Xception} & 0.852 $\pm$ 0.018 & 0.948 $\pm$ 0.008 & \textbf{0.952 $\pm$ 0.003} \\ 
\textbf{Inception} & 0.888 $\pm$ 0.018 & 0.905 $\pm$ 0.029 & \textbf{0.947 $\pm$ 0.010} \\ 
\bottomrule
\end{tabular}
\end{table}

In order to validate our findings, we repeated our experiments for iris PAD task.
For RQ1, we saw similar findings as for synthetic face detection, as three out of the four student model architectures benefited from AI Teachers taught using human saliency. For RQ2, the top-performing AI Teachers improved the performance of AI Students across different CNN architectures. Finally, we were able to increase the performance of AI Student models over human-guided teacher models by selecting optimal teacher saliency and alpha values (RQ3), as shown in Tab. \ref{tab:iris-optimal}. We included non-essential, yet potentially informative results and graphs related to the iris PAD results in the supplementary materials.

\section{Conclusions}

We have proposed, designed and evaluated a learning framework that makes an efficient use of limited human saliency data, allowing to significantly scale human-guided training strategies. To accomplish this goal, we first use a small amount of human annotations to train AI Teachers, that is, models that generate saliency for subsequent AI Students. These student models are trained using existing saliency-guided training paradigms, but utilizing synthetically-generated salience rather than human-supplied salience. 

We extensively tested our framework in a task of synthetic face detection, and explored selected variants in a task of iris presentation attach detection (to check the domain generalization hypothesis). We observed a boosted performance of the resulting student models trained by AI Teachers built using human salience, when compared to student models trained without any salience information. Even more importantly, we also saw a better performance when models trained with human salience were used as AI Teachers, compared to teacher models not exposed to human salience before. That confirms the usefulness of incorporating human salience into CNN training, and this paper -- to our knowledge -- for the first time demonstrates how to leverage small availability of human annotations and scale the human perception-augmented training. The proposed way of learning can thus serve as one of the ideas to match the growing size of datasets in any domain in which humans can provide initial limited salience information sufficient to train teacher models. 

\bibliography{bmvc_final}

\end{document}

\title{Teaching AI to Teach: \\Leveraging Limited Human Salience Data Into Unlimited Saliency-Based Training\\(Supplementary Materials)}

\addauthor{Colton R. Crum}{ccrum@nd.edu}{1}
\addauthor{Aidan Boyd}{aboyd3@nd.edu}{1}
\addauthor{Kevin Bowyer}{kwb@nd.edu}{1}
\addauthor{Adam Czajka}{aczajka@nd.edu}{1}

\addinstitution{
 University of Notre Dame\\
 Notre Dame, IN 46556, USA
}
\runninghead{Crum, Boyd, Bowyer, Czajka}{Teaching AI to Teach}

\def\eg{\emph{e.g}\bmvaOneDot}
\def\ie{\emph{i.e}\bmvaOneDot}
\def\Eg{\emph{E.g}\bmvaOneDot}
\def\etal{\emph{et al}\bmvaOneDot}

\maketitle

 \renewcommand{\thefigure}{S\arabic{figure}}
 \renewcommand{\thetable}{S\arabic{table}}
 
\noindent
These supplementary materials illustrate selected aspects of the presented work. 

\begin{figure}[!htb]
    
    \centering
    \includegraphics[width=\linewidth]{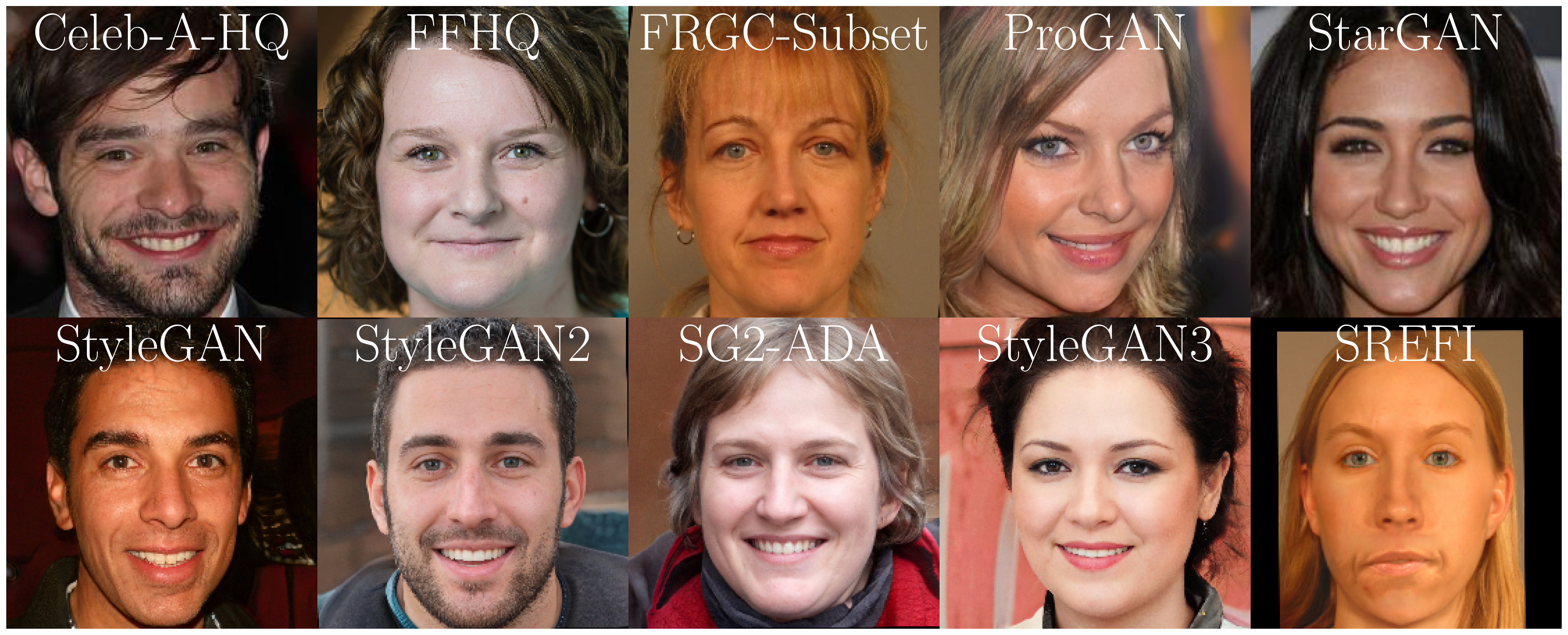}
    \vskip2mm
    \caption{Examples of the \textbf{authentic} (Celeb-A-HQ, FFHQ, FRGC-Subset) and \textbf{synthetic} (ProGAN, StarGAN, StyleGAN, StyleGAN2, StyleGAN2-ADA, StyleGAN3, SREFI) samples from the dataset used to train the AI Students.}
    \label{fig:face-data}
\end{figure}

\begin{figure}[!htb]
    \centering
    \includegraphics[width=\linewidth]{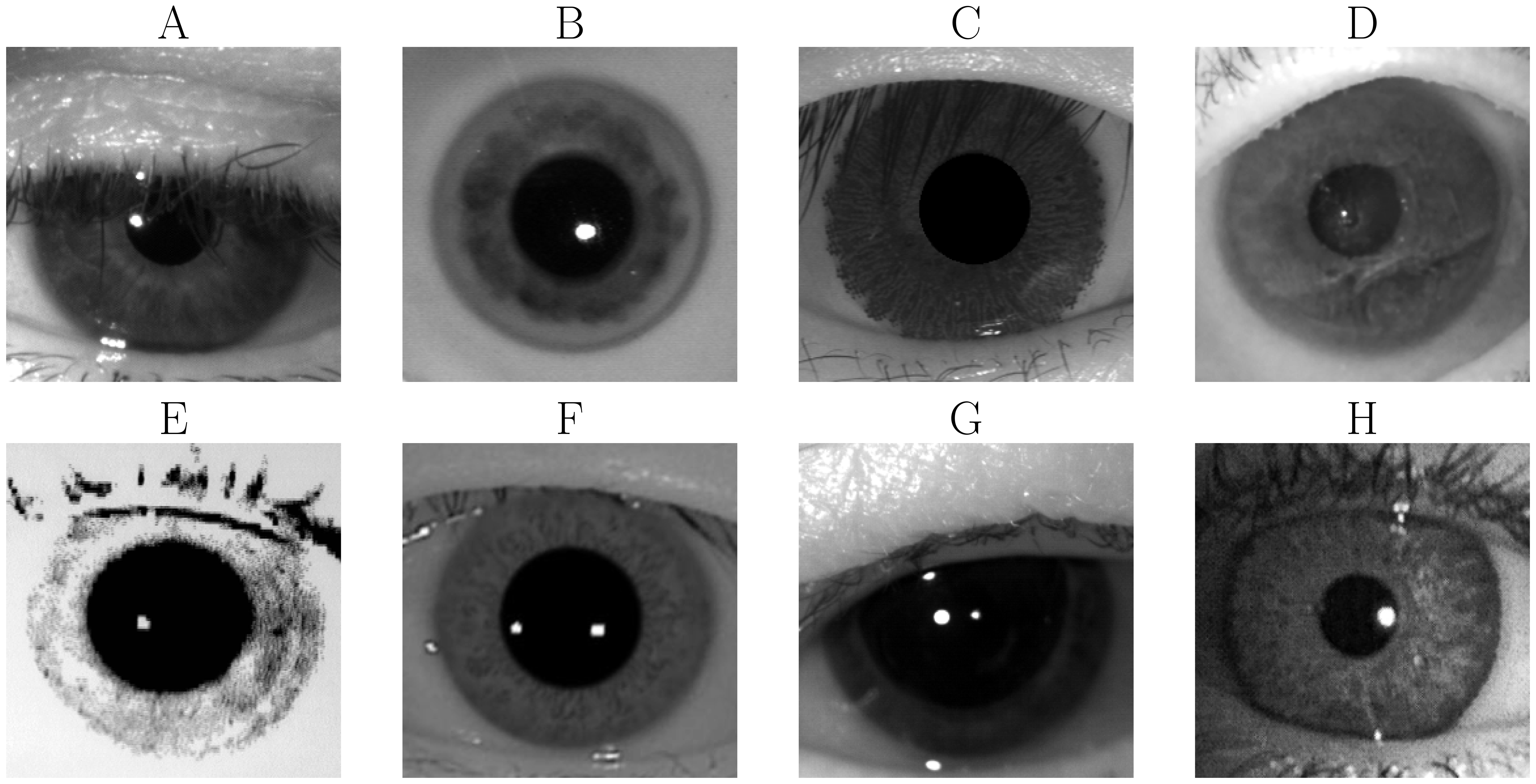}
    \vskip2mm
    \caption{Examples of \textbf{live sample} (A) and \textbf{spoof samples:} aratifical (B), textured contact lenses (C), post-mortem (D), paper printouts (E), synthetically-generated (F), diseased (G), and wearing contacts then printed (H). Samples taken from the dataset used to train the AI Students.}
    \label{fig:iris-data}
\end{figure}

\vskip2mm\noindent
Figures \ref{fig:face-data} and \ref{fig:iris-data} show authentic and synthetically-generated (or faked in other ways) face and iris examples, respectively, taken from datasets used in this work. 

\vskip2mm\noindent
Table \ref{tab:iris-results} mimics Table 1 in the main paper, except that we present results for the task of iris presentation attack detection.

\vskip2mm\noindent
Table \ref{tab:amount} shows the performance impact of AI Students after reducing the number of human annotations used to train AI Teachers for iris PAD.

\begin{figure}[!htb]
    \centering
      \begin{subfigure}[b]{0.263\linewidth}
          \centering
          \includegraphics[width=\linewidth]{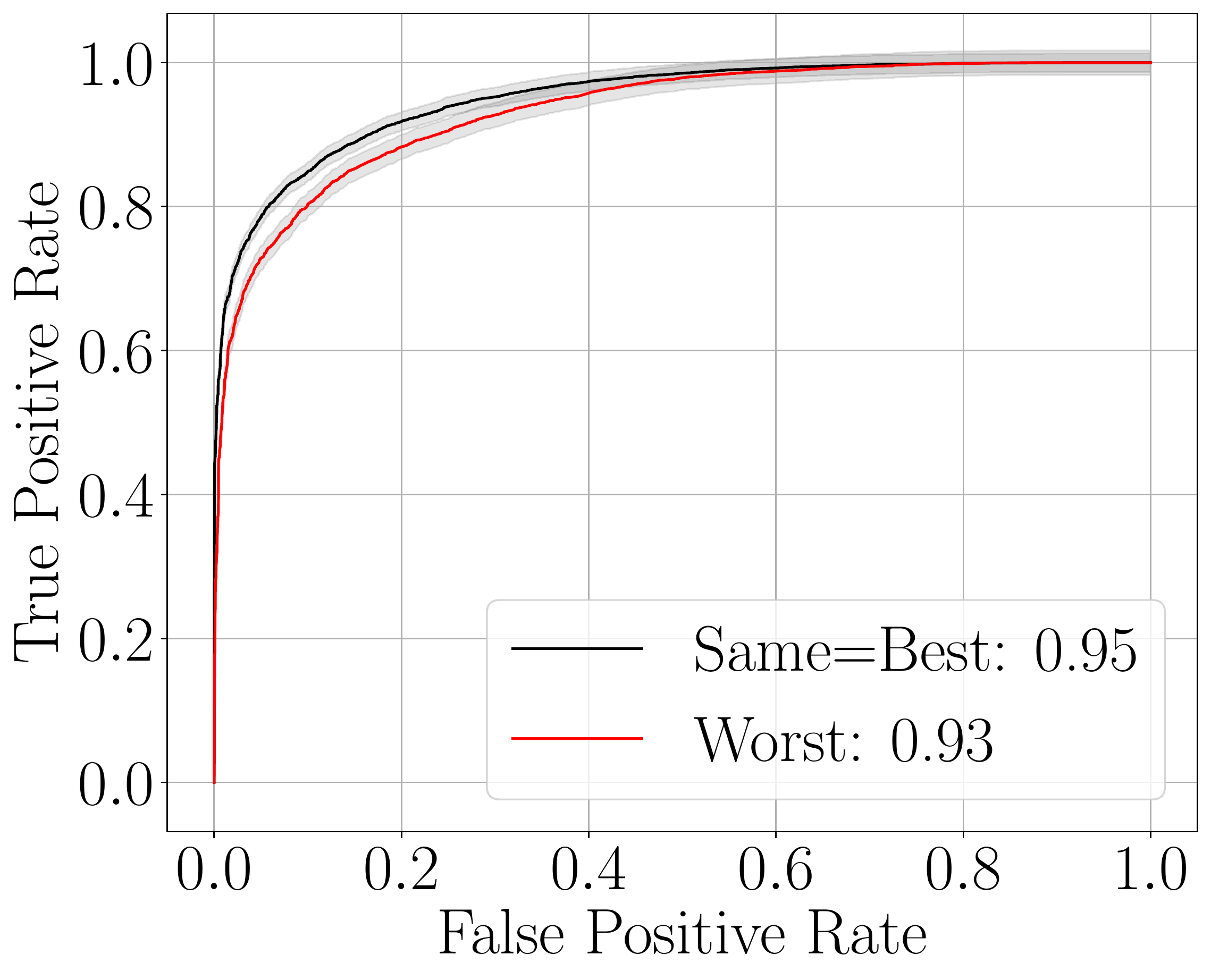}
          \caption{DenseNet}
      \end{subfigure}\hfill
      \begin{subfigure}[b]{0.23\linewidth}
          \centering
          \includegraphics[width=\linewidth]{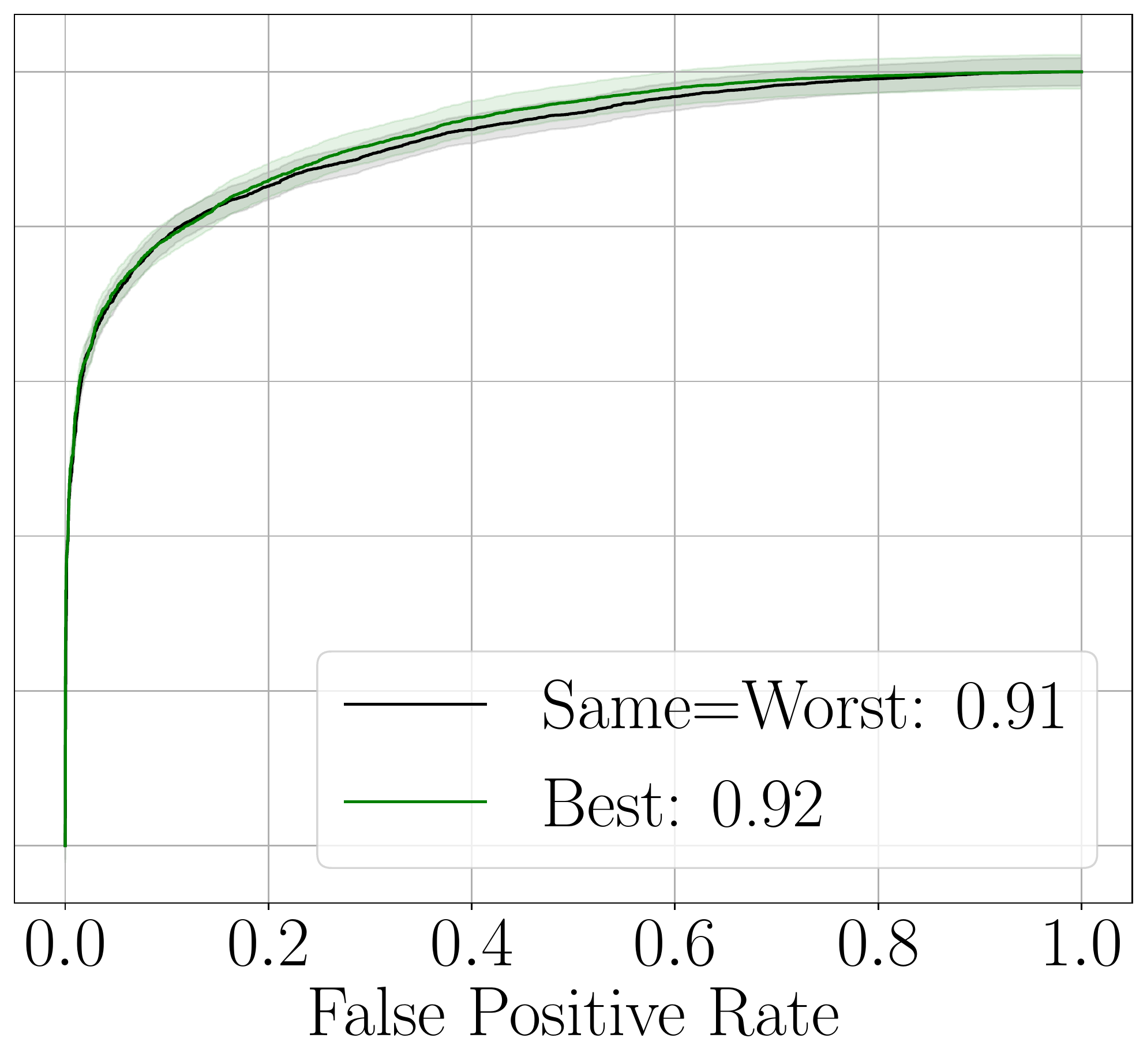}
          \caption{ResNet}
      \end{subfigure}\hfill
      \begin{subfigure}[b]{0.23\linewidth}
          \centering
          \includegraphics[width=\linewidth]{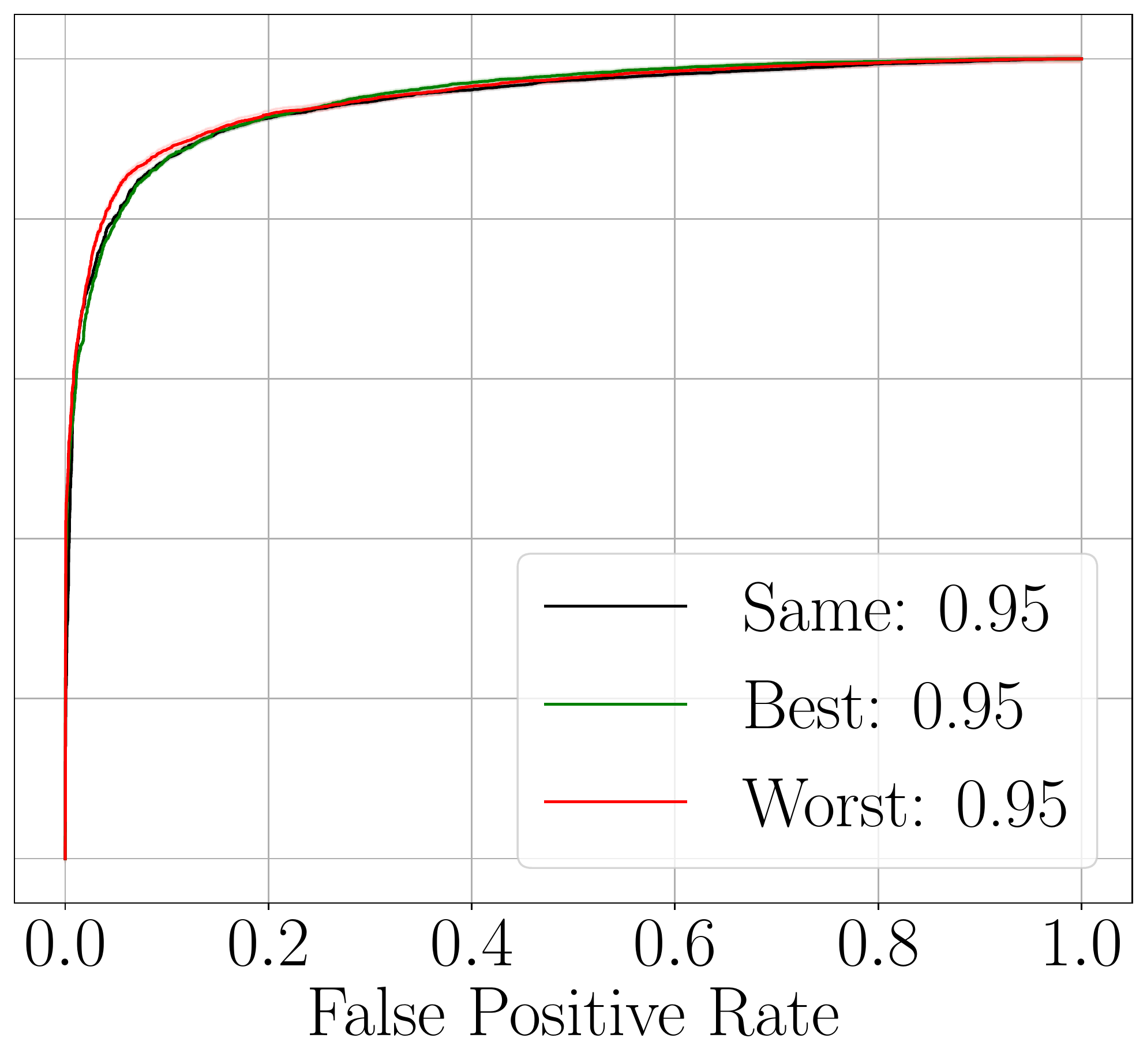}
          \caption{Xception}
      \end{subfigure}\hfill
      \begin{subfigure}[b]{0.23\linewidth}
          \centering
          \includegraphics[width=\linewidth]{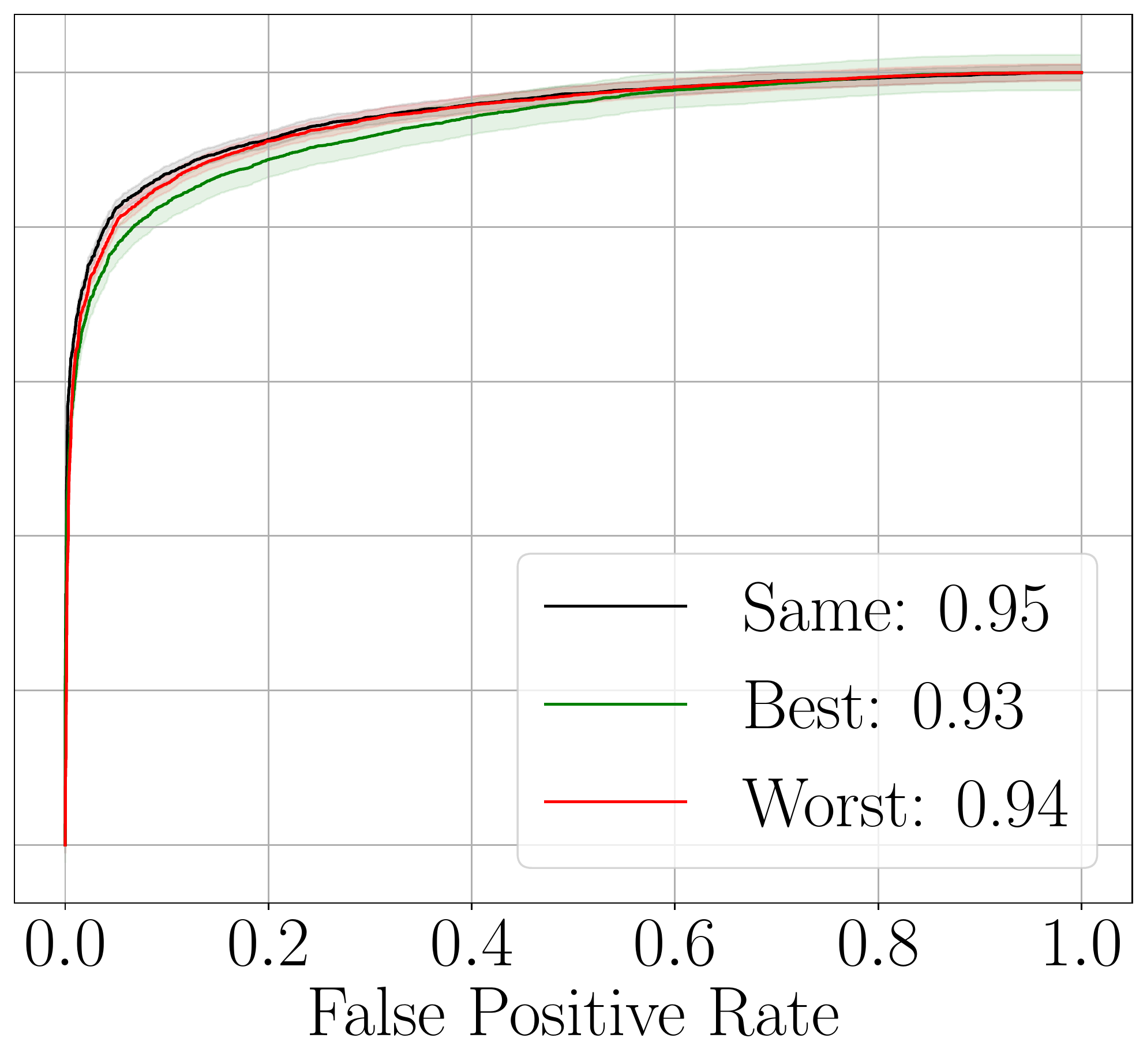}
          \caption{Inception}
      \end{subfigure}
    \caption{Same as in Figure 3 in the main paper, except that results for iris PAD are shown.}
    \label{fig:iris-best-worst-experiment}
\end{figure}

\vskip2mm\noindent
Figure \ref{fig:iris-best-worst-experiment} mimics Figure 3 in the main paper, except that we present results for the task of iris presentation attack detection.

\begin{figure}
  \centering
  \includegraphics[width=250pt]{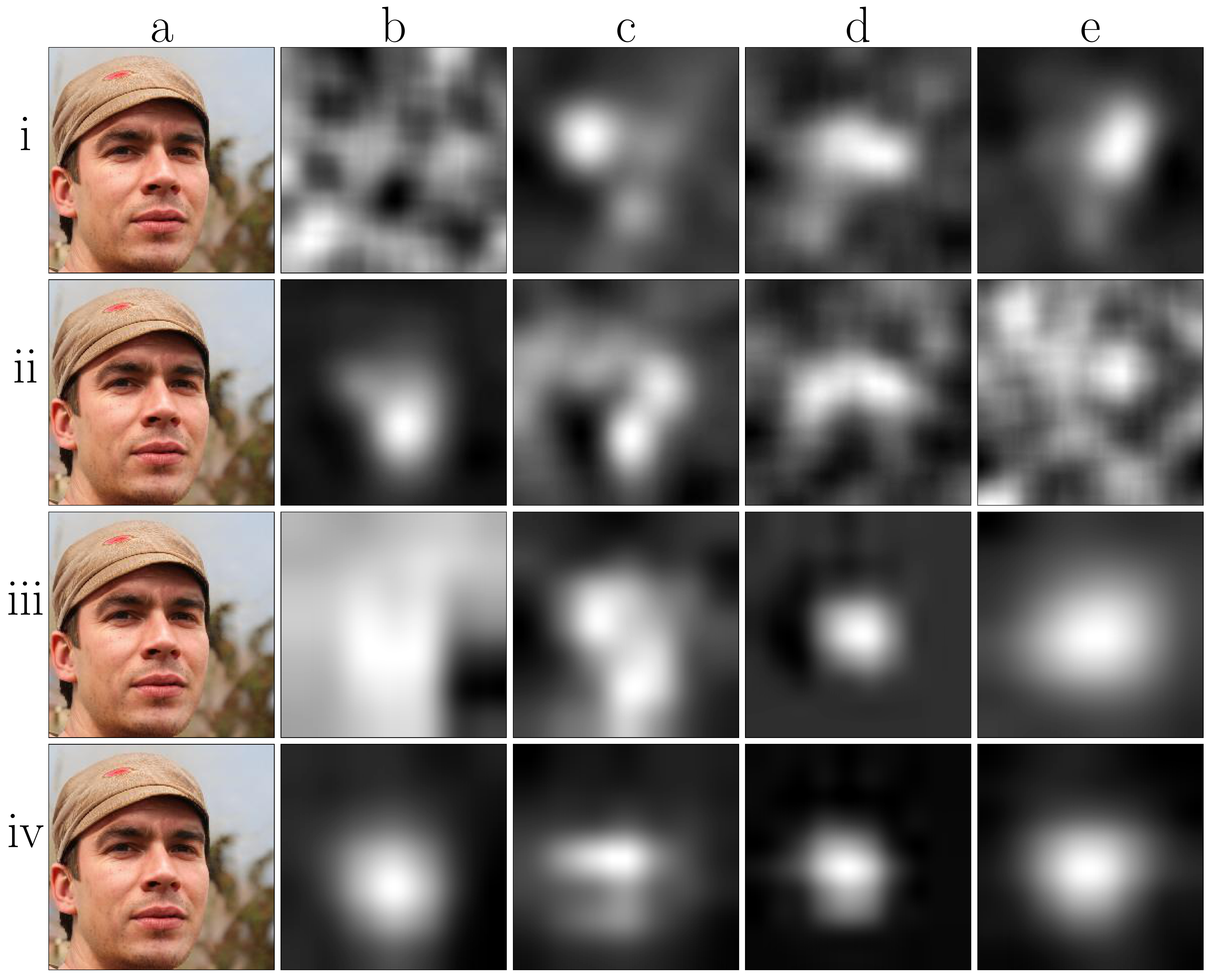}
  \vskip2mm
  \caption{Illustration of AI Teacher model's salience maps on the TAIT training data for {\bf synthetic face detection}: (a) input sample, (b) DenseNet, (c) ResNet, (d) Xception, (e) Inception; (i) RISE-based salience + cross-entropy training, (ii) RISE-based salience + human-guided training, (iii) CAM-based salience + cross-entropy training, (iv) CAM-based salience + human-guided training.}
  \label{fig:proxy_example_face}
\end{figure}

\begin{figure}
  \centering
  \includegraphics[width=250pt]{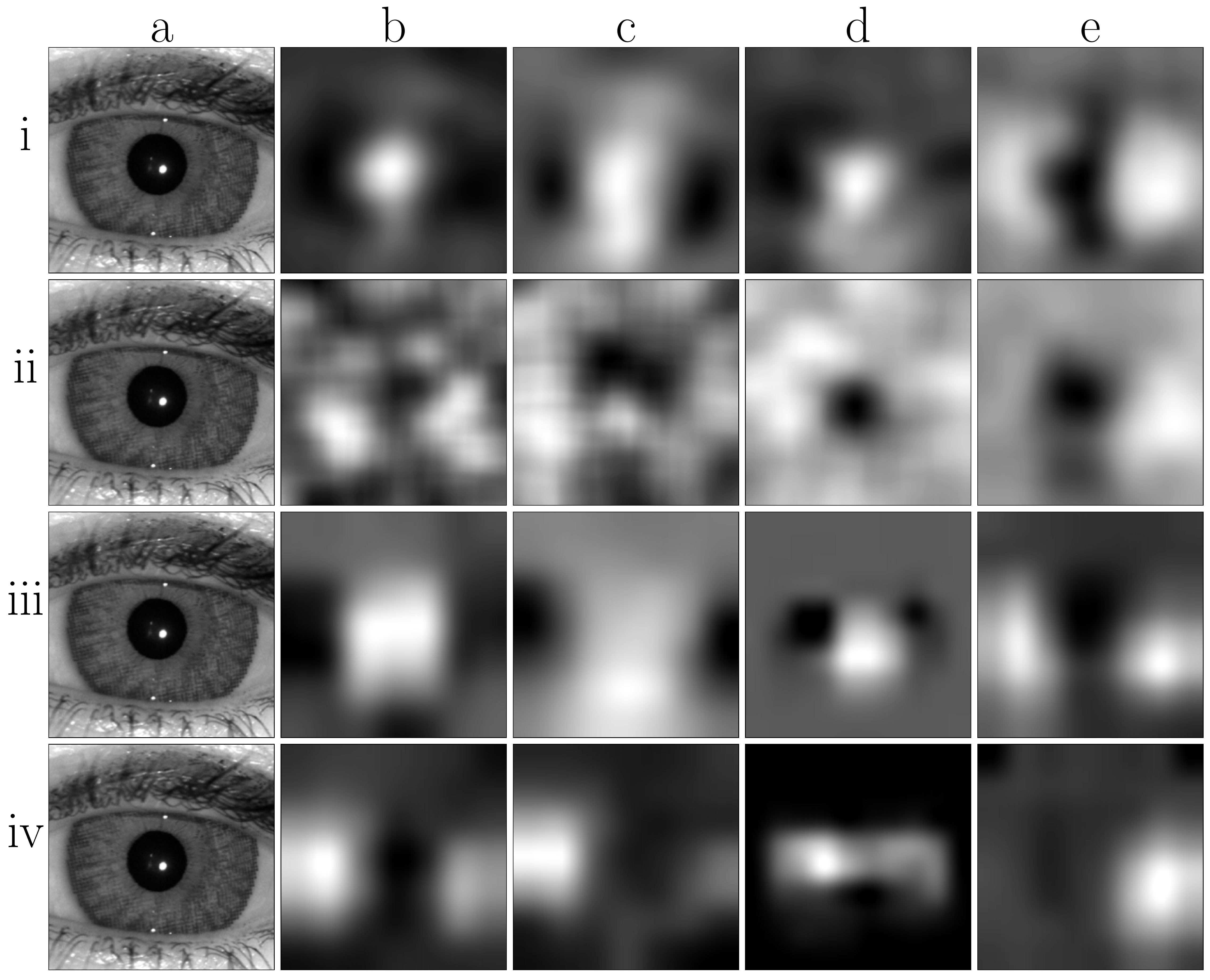}
  \vskip2mm
  \caption{Illustration of AI Teacher model's salience maps on the TAIT training data for {\bf iris PAD detection}: (a) input sample, (b) DenseNet, (c) ResNet, (d) Xception, (e) Inception; (i) RISE-based salience + cross-entropy training, (ii) RISE-based salience + human-guided training, (iii) CAM-based salience + cross-entropy training, (iv) CAM-based salience + human-guided training.}
  \label{fig:proxy_example_iris}
\end{figure}

\vskip2mm\noindent
Finally, Figures \ref{fig:proxy_example_face} and \ref{fig:proxy_example_iris} show examples of AI Teachers-generated saliency for models trained in various ways (with and without human saliency), for two types of saliency generation (CAM and RISE) and across various architectures.

\begin{table}[!htb]
\centering
\caption{Same as in Table 1 in the main paper, except that results for iris PAD are shown.}

\vskip2mm

\begingroup
\setlength{\tabcolsep}{2pt}
\renewcommand{\arraystretch}{1}
\small
\begin{tabular}{cccccc}
\toprule
{\bf How the AI Teachers} & \multicolumn{4}{c}{\bf Mean AUC on the EAIS data} \\\cline{2-5}
{\bf were trained on TAIT data}  & {\bf DenseNet} & {\bf ResNet} & {\bf Xception} & {\bf Inception}\\

\midrule
 
 Without human salience & \textcolor{CAM}{0.944}\textcolor{CAM}{$\pm$}\textcolor{CAM}{0.009} & \textcolor{CAM}{0.908}\textcolor{CAM}{$\pm$}\textcolor{CAM}{0.042} & \textcolor{RISE}{\textbf{0.952}}\textcolor{RISE}{$\pm$}\textcolor{RISE}{\textbf{0.003}} & \textcolor{RISE}{0.939}\textcolor{RISE}{$\pm$}\textcolor{RISE}{0.015} \\
  \cline{1-5}
  With human salience & \textcolor{CAM}{\textbf{0.950}}\textcolor{CAM}{$\pm$}\textcolor{CAM}{\textbf{0.013}} & \textcolor{CAM}{\textbf{0.915}}\textcolor{CAM}{$\pm$}\textcolor{CAM}{\textbf{0.018}} & \textcolor{RISE}{0.950}\textcolor{RISE}{$\pm$}\textcolor{RISE}{0.003}  & \textcolor{CAM}{\textbf{0.947}}\textcolor{CAM}{$\pm$}\textcolor{CAM}{\textbf{0.010}} \\
 \bottomrule
 \end{tabular}
 \endgroup
 \label{tab:iris-results}
 \end{table}


\begin{table}[t!]
\label{tab:amount}
\centering
\caption{AUC for iris PAD using DenseNet as the backbone for Teachers and Students. Same training configurations as the main paper, except different amounts of data were used to train Teachers. The number of samples are split evenly between both classes. The table shows a single train-test experiment.}
\vskip3mm
\footnotesize
\begin{tabular}{c|c}
\multirow{2}{*}{\parbox{5cm}{\centering\textbf{\# of human annotated samples\\ used to train Teacher models}}}&\multirow{2}{*}{\parbox{1.8cm}{\centering\textbf{Student \\performance}}}\\
    &  \\ \hline
100 (13\% of human annotated set) & 0.905 \\ \hline
382 (50\% of human annotated set) & 0.940 \\
\end{tabular}
\label{tab:amount}
\vskip3mm
\end{table}